# A Parameter-Efficient Transfer Learning Approach through Multitask Prompt Distillation and Decomposition for Clinical NLP

**Cheng Peng, PhD[1], Mengxian Lyu, MS[1], Ziyi Chen, MS[1], Yonghui Wu, PhD[1,2]**

**[1]Department of Health Outcomes and Biomedical Informatics, College of Medicine, University of Florida, Gainesville, Florida, USA; [2]Cancer Informatics Shared Resource, University of Florida Health Cancer Center, Gainesville, Florida, USA**

**Abstract**

*Existing prompt-based fine-tuning methods typically learn task-specific prompts independently, imposing significant computing and storage overhead at scale when deploying multiple clinical natural language processing (NLP) systems. We present a multitask prompt distillation and decomposition framework that learns a single shared meta-prompt from 21 diverse clinical source tasks and adapts it to unseen target tasks with fewer than 0.05% trainable parameters. Evaluated across five clinical NLP task types (named entity recognition, relation extraction, question answering, natural language inference, and summarization) on 10 held-out target datasets using three backbone models (LLaMA 3.1 8B, Meditron3 8B, gpt-oss 20B), our framework consistently outperforms LoRA by 1.5-1.7% despite using orders of magnitude fewer parameters, and exceeds single-task prompt tuning by 6.1–6.6%. The gpt-oss 20B model achieves the highest overall performance, particularly on clinical reasoning tasks. The strong zero- and few-shot performance demonstrates better transferability of the shared prompt representation.*

## Introduction

Large Language Models (LLMs) have revolutionized Clinical Natural Language Processing (NLP), achieving near-human performance on complex tasks[1,2] ranging from information extraction (e.g., clinical concept and relation extraction) to clinical reasoning (e.g., question answering and summarization). In the current human-machine interaction, humans use prompts, i.e., instructions in human language, to instruct LLMs to perform specific tasks. However, integrating these capabilities into routine hospital workflows poses a persistent bottleneck in handling instructions across multiple tasks, which jeopardizes the adoption of LLMs. First, models trained for one task type do not transfer to other task types. The current clinical information extraction system cannot leverage instructions from other tasks, such as natural language inference (NLI) or summarization, despite operating on the same underlying clinical text. Second, models trained at one institution routinely fail when deployed at another institution due to systematic variation in documentation culture, EHR system differences, local clinical vocabularies, and patient population demographics[3]. Third, models trained on one disease domain generalize poorly to others. A system optimized for oncology notes may fail to capture cardiology documentation due to differences in clinical terminology and the distribution of relevant concepts[4]. To overcome these generalization failures, modern healthcare systems often develop multiple highly specialized models for each sub-domain application. However, traditional full-model fine-tuning requires massive amounts of annotated data, which is scarce in clinical settings due to the high cost of expert annotation and strict privacy regulations governing access to clinical text. As recent LLM sizes scale into the billions of parameters, updating and storing a separately fine-tuned model for each task, institution, and disease domain becomes computationally prohibitive and practically unsustainable for resource-constrained hospital IT infrastructures.

Parameter-efficient fine-tuning (PEFT) methods have been increasingly adopted to mitigate deployment costs by freezing the backbone and updating only a small fraction of the model's parameters.[5] Adapter methods[5] insert lightweight bottleneck modules between transformer layers, achieving strong performance with roughly 3–4% additional parameters per task. Low-rank adaptation (LoRA)[6] decomposes weight update matrices into low-rank factors, matching full fine-tuning performance on many benchmarks with approximately 1–2% of trainable parameters and no inference overhead. Soft prompt tuning[7] and prefix tuning[8] take a different approach, prepending a small set of learnable continuous vectors to the model input while keeping all backbone parameters frozen. At scale, soft prompt tuning approaches full fine-tuning performance while updating fewer than 0.1% of total parameters, making it the most parameter-efficient adaptation strategy. However, soft prompt tuning has well-documented limitations: it is unstable for smaller models, and it learns each task prompt from scratch, failing to exploit the shared structure and transferable knowledge across related clinical tasks.

A more recent solution to enhance the stability and generalizability of single-task prompt tuning is to initialize target-task prompts from prompts learned on related source tasks. SPoT[9] retrieves the most similar source task prompt as initialization for the target task, improving over random initialization but limited to a single source. ATTEMPT[10] uses an attention mechanism over a pool of source prompts, allowing soft interpolation across multiple sources, which is more flexible but adds inference-time overhead with interpolated prompt vectors. Multitask Prompt Tuning (MPT)[11] takes a fundamentally different approach: it learns a single shared meta-prompt matrix by decomposing each task prompt, then distilling knowledge from independently trained teacher prompts into this shared representation. MPT achieves state-of-the-art prompt transfer performance while using only 0.035% of tunable parameters per target task, significantly outperforming both other PEFT methods and transferable prompting methods in parameter efficiency and matching or exceeding full fine-tuning on several benchmarks. While MPT performed well on general NLP benchmarks, it has not been rigorously evaluated in the clinical domain. It remains unknown whether a shared, transferable prompt learned from diverse clinical source tasks can generalize to unseen clinical target tasks, whether clinical domain pretraining improves the quality and transferability of the shared prompt representation, and how multitask soft prompt transfer compares with other PEFT methods, such as LoRA, in the low-resource clinical adaptation setting.

This study addresses these challenges through a comprehensive empirical study of multitask prompt tuning for transfer learning in clinical NLP. Our contributions are as follows. (1) We present a prompt distillation and decomposition framework, where a task-level teacher prompt training strategy with unified label spaces enables knowledge distillation and prompt decomposition across heterogeneous clinical datasets within each task type. (2) We construct a multitask clinical NLP benchmark dataset, comprising 21 source datasets across five clinical NLP tasks, including named entity recognition (NER), relation extraction (RE), question answering (QA), natural language inference (NLI), and text summarization. (3) We provide a comprehensive empirical evaluation using three state-of-the-art generative LLMs, including the general-domain models (LLaMA 3.1 8B[12] and gpt-oss 20B[13]) and a clinical-domain model (Meditron3 8B[14]), and evaluate their efficacy on 10 held-out clinical target tasks spanning cross-task, cross-institutional, and cross-disease transfer scenarios. (4) We conduct systematic few-shot learning evaluations to characterize MPT's behavior in the low-resource regime that defines most real-world clinical NLP deployment scenarios. Our results demonstrate that a single shared prompt learned from diverse clinical source tasks achieve comparable performance to full fine-tuning and consistently outperforms task-specific prompt initialization across target tasks, that Meditron3's clinical pretraining yields stronger and more transferable shared prompt representations than the matched general-domain backbone, and that MPT achieves competitive or superior performance to LoRA with an order of magnitude fewer tunable parameters, establishing multitask prompt distillation and decomposition as a practical and principled approach to scalable clinical NLP deployment.

## Methods

### Task Formulation

We formulate clinical NLP transfer learning as a multitask prompt transfer problem. Let $S = \{S_1, S_2, S_3, S_4, S_5\}$ denote five clinical source task types (NER, RE, QA, NLI, and summarization), where each Sk aggregates multiple heterogeneous clinical datasets sharing a common task structure and unified label space. Let $T = \{T_1, \ldots, T_{10}\}$ denote ten held-out clinical target tasks drawn from the same five task types but from distinct datasets, institutions, and disease domains unseen during source training. Given a frozen pretrained LLM with parameters $\Theta$, our objective is to learn a single shared meta-prompt matrix $P^* \in \mathbb{R}^{L \times d}$ from $S$, where $L$ is the prompt length, and $d$ is the model hidden dimension, such that $P^*$ can be efficiently adapted to any target task $T_t$ by updating only a minimal set of task-specific parameters while $\Theta$ remaining frozen throughout. All five task types are reformulated under a unified text generation framework, where NER and RE outputs are generated as structured entity and relation sequences, NLI outputs are generated as label tokens, and QA and summarization outputs are generated as free-form text. This unified formulation enables a single training and inference pipeline across all task types without task-specific output heads.

### Multitask Prompt Transfer Learning Framework

**Task-level teacher prompt training.** For each of the five clinical source tasks, we unify label spaces across heterogeneous datasets and train a single teacher prompt on the combined training data. We initialize a continuous soft prompt $P_k \in \mathbb{R}^{L \times d}$, where $L$ denotes the prompt length (number of virtual tokens) and $d$ denotes the embedding dimension. Keeping the LLM backbone entirely frozen, we independently train each teacher prompt $P_k$ on its respective unified task corpus using standard cross-entropy loss over the generated text. This yields five robust, task-specific teacher prompts.

**Multitask prompt distillation and decomposition.** To extract the transferable clinical knowledge shared across all tasks, we decompose the independently learned teacher prompts into a single shared meta-prompt and task-specific low-rank updates. We define the shared meta-prompt as $P^*$.The reconstructed prompt for task $k$, denoted as $\hat{P}_k$, is mathematically formulated using the Hadamard product (element-wise multiplication):

$$\hat{P}_k = P^* \odot (U_k \times V_k)$$

where $U_k \in \mathbb{R}^{L \times r}$ and $V_k \in \mathbb{R}^{r \times d}$ are small trainable matrices for each source task $k$, and $r$ is the bottleneck rank such that $r \ll \min(L, d)$. $P^*$ and all task-specific vectors are trained jointly by distilling knowledge from the teacher prompts via a three-component loss: (1) a task loss $\mathcal{L}_{\text{task}}$ on target outputs using the decomposed prompt; (2) a logit distillation loss $\mathcal{L}_{\text{logits}}$ as KL divergence between teacher and student output distributions; and (3) a hidden state distillation loss $\mathcal{L}_{\text{hidden}}$ computed as MSE between teacher and student final-layer hidden states, encouraging representational alignment beyond surface output matching. The total loss per task is:

$$\mathcal{L}_k = \mathcal{L}_{\text{task}} + \lambda_1 \mathcal{L}_{\text{logits}} + \lambda_2 \mathcal{L}_{\text{hidden}}$$

where $\mathcal{L}_{\text{task}} = \sum_{k \in S} \mathcal{L}_{task}^{k}$ represents the aggregated task losses for all source tasks, and $\lambda_1, \lambda_2$ are weights to balance the impact of distillation loss terms.

**Target task adaptation.** During the evaluation phase on the unseen target datasets, the learned $P^*$ is adapted to each held-out target task by initializing the target prompt as:

$$\hat{P}_t = P^* \odot (u_t \otimes {v_t}^T)$$

where $u_t$ and $v_t$ are initialized from the source task-specific vectors of the matching task type. For example, a target NER task is initialized from the NER source vectors. During adaptation, $P^*$ is kept frozen and only $u_t$ and $v_t$ are updated. Each task contains the shared prompt $L \times d$ that has the same dimensions as a vanilla soft prompt and a smaller number of task-specific vectors $(L + d)$. Thus, the total number of tunable parameters for a single target task is $(L \times d) + (L + d)$. After training, this can further be compressed into a single matrix of size $L \times d$. For a group of target tasks, the total number of tunable parameters is $(L \times d) + (L + d)\tau$ , where $\tau$ is the number of target tasks.

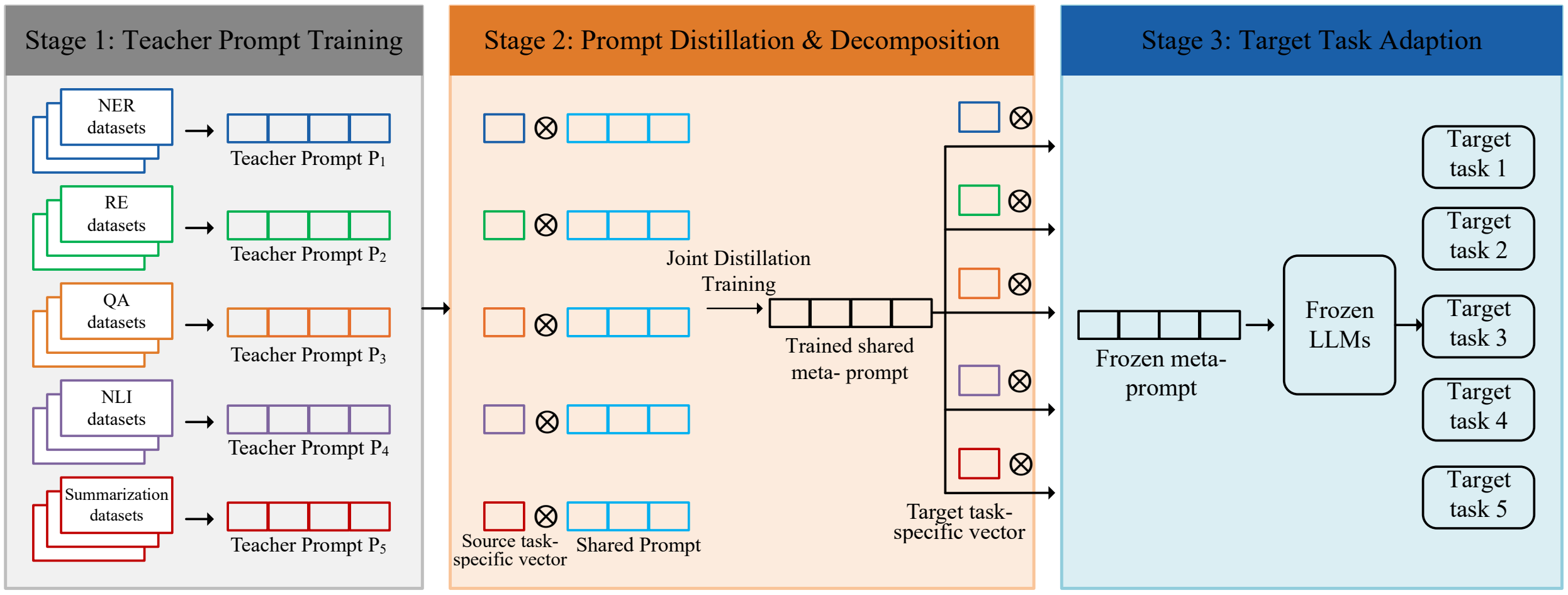

**Figure 1**. Overview of the multitask prompt distillation and decomposition framework. Stage 1 trains independent teacher prompts across five clinical NLP source tasks. Stage 2 decomposes the teacher prompts into a shared meta-prompt and task-specific low-rank factors through joint distillation training. Stage 3 adapts the frozen shared meta-prompt to unseen target tasks by fine-tuning only task-specific vectors.

## Tasks and datasets

We evaluate our framework across five clinical NLP task types using 21 source datasets and 10 target datasets. Table 1 summarizes the composition of source and target datasets, including task type, dataset name, sample size, and domain coverage.

**Source datasets** span all five task types. For NER, we use the 2018 and 2022 n2c2 shared task[15,16] corpora covering medication, adverse event entities and social determinants of health (SDoH) from MIMIC-III clinical notes, BC5CDR[17] and NCBI Disease[18] covering chemical and disease entities from biomedical abstracts, and an institutional oncology corpus from UFHealth. For RE, we use the 2018 and 2022 n2c2 corpora for drug-event and SDoH relation extraction, ChemProt[19] for chemical-protein interactions, BC5CDR for chemical-disease relations, and the UFHealth oncology corpus. For QA, we use MedQA[20] (USMLE-style multiple choice), MedMCQA[21] (Indian medical licensing), PubMedQA[22] (biomedical QA), and emrQA[23] (extractive QA over clinical notes). For NLI, we use MedNLI[24] (clinical note inference), BioNLI[25] (biomedical literature inference), SDOH-NLI[26] (social determinants of health), and NLI4CT[27] (clinical trial inference). For summarization, we use MIMIC-CXR[28] (radiology report impression generation), PubMed-summarization[29] (biomedical abstract summarization), and CHQ-Summ[30] (consumer health question summarization).

**Target datasets** are selected to evaluate three distinct transfer scenarios. Cross-institutional transfer is evaluated using the 2022 n2c2 University of Washington split for NER and RE, where source training uses only the MIMIC-III split, testing generalization across hospital systems. Cross-disease transfer is evaluated using the UFHealth Opioid NER and RE corpora, reflecting a shift from the oncology and general medication domains seen in the source training. Cross-domain transfer is evaluated using HEAD-QA[31] (medical licensing QA), Medbullets[32] (clinical vignette QA), SciNLI[33] (scientific NLI), RadNLI[34] (radiology NLI), RadSum[35] (radiology summarization), and MeQSum[36] (consumer health summarization).

**Table 1.** Summary of source and target datasets used for multitask prompt training and evaluation.

| | Task type | Dataset | Samples | Domain |
|---|---|---|---|---|
| Source Tasks and Datasets | NER | 2018 n2c2, 2022 n2c2 (MIMIC-III), BC5CDR, NCBI, UFHealth Cancer | ~47,000 | Clinical/Biomedical |
| | RE | 2018 n2c2, 2022 n2c2, ChemProt, BC5CDR, UFHealth Cancer | ~41,000 | Clinical/Biomedical |
| | QA | MedQA, MedMCQA, PubmedQA, emrQA | ~193,500 | Clinical/Biomedical |
| | NLI | MedNLI, BioNLI, SDOH-NLI, NLI4CT | ~24,700 | Clinical/Biomedical |
| | Summarization | MIMIC-CXR, PubMed-summarization, CHQ-Summ | ~348,400 | Clinical/Biomedical |
| Target Tasks and Datasets | NER | 2022 n2c2 (UW), UFHealth Opioid use | ~3,500 | Clinical |
| | RE | 2022 n2c2 (UW), UFHealth Opioid use | ~3,500 | Clinical |
| | QA | HEAD-QA, Medbullets | ~2,800 | Clinical/Biomedical |
| | NLI | SciNLI, RadNLI | ~4,500 | Clinical/Biomedical |
| | Summarization | RadSum, MeQSum | ~4,000 | Clinical/Biomedical |

## Experimental setup

**Models and baselines.** We evaluate our framework on three widely used generative LLMs across four adaptation methods:

LLaMA 3.1 8B: A state-of-the-art general-domain model, providing a baseline for broad linguistic reasoning without specialized medical alignment.

Meditron3 8B: A clinical-domain model initialized from the LLaMA 3.1 architecture and continuously pre-trained on a comprehensive corpus of biomedical literature and clinical guidelines.

gpt-oss 20B: A general-domain Mixture-of-Experts (MoE) architecture. Including this sparse model allows us to investigate the impact of model scale and advanced architecture during prompt transfer.

Baselines include Full Fine-Tuning (Full FT, 100% params), LoRA (rank r = 16, ~2.50% params), single-task Prompt Tuning (PT, random initialization, <0.05% params), and our proposed MPT (<0.05% params).

**Implementation details.** All soft prompt methods use a prompt length of $L = 100$ tokens initialized from randomly sampled vocabulary embeddings. In Stage 1, each teacher prompt is trained for 5 epochs using AdamW with learning rate 0.05, and batch size 32, with large datasets subsampled to 50K examples. In Stage 2, $P^*$ and all task-specific vectors are trained jointly for 10 epochs with learning rate 0.01, and loss weights $\lambda_1 = \lambda_2 = 0.5$. Stochastic task sampling draws $K \in \{2, 3, 4, 5\}$ task types uniformly per training step to prevent corpus-size domination. In Stage 3, target task-specific vectors are fine-tuned for up to 10 epochs with early stopping on the validation loss at a learning rate of 0.01, with the few-shot settings using a fixed 50 steps. All experiments are conducted on 4 NVIDIA B200 192GB GPUs.

**Evaluation metrics.** NER and RE are evaluated using the micro-averaged F1 score. QA uses accuracy for multiple-choice datasets and F1 for extractive settings. NLI uses macro-averaged accuracy. Summarization uses the ROUGE-L score. Few-shot results report mean ±standard deviation across 10 random draws of the k-shot training set at $k \in \{0, 1, 5, 10, 20\}$.

## Results

Table 2 shows results across 10 clinical tasks under full dataset adaptation. Across all three generative LLMs, MPT consistently outperforms LoRA despite using substantially fewer trainable parameters (<0.05% vs ~2.50%), with average margins of 1.7%, 1.6%, and 1.5% over LLaMA 3.1 8B, Meditron3 8B, and gpt-oss 20B, respectively. The MPT-LoRA gap is largest on RE and cross-disease transfer tasks (e.g., RE-Opioid: +2.2% for LLaMA, +2.1% for Meditron3) and smallest on NER and NLI tasks, where task structure provides a stronger adaptation signal even for LoRA. MPT also outperforms single-task prompt tuning (PT) by a substantially larger margin across all models and tasks, with average improvements of 6.6%, 6.4%, and 6.1% respectively. The MPT-PT gap is most remarkable on cross-institutional NER and RE tasks (e.g., NER-UW: +7.2% for LLaMA), while the gap is smaller on QA tasks (e.g., HEAD-QA: +6.1% for LLaMA). Among the two 8B models, Meditron3 8B outperforms LLaMA 3.1 8B across all methods by an average margin of 2.3%, with the largest averaged gains of 2.8% on NER and 2.5% on RE, and more modest gains on summarization tasks (1.5-1.8%). Notably, Meditron3 8B MPT (avg. 0.715) exceeds LLaMA 3.1 8B Full FT (avg. 0.699) while using less than 0.05% of its parameters. gpt-oss 20B achieves the highest performance across all tasks and methods, with the largest margins over Meditron3 8B on QA tasks (HEAD-QA: +4.6%, Medbullets: +4.6% over Meditron3 8B Full FT) where model scale and MoE architecture contribute most to clinical reasoning, and more modest margins on structured prediction tasks (NER-UW: +1.5%, RE-UW: +1.8%). gpt-oss 20B MPT (avg. 0.739) falls within 0.7% of gpt-oss 20B Full FT (avg. 0.746) and outperforms both LLaMA 3.1 8B Full FT (avg. 0.699) and Meditron3 8B Full FT (avg. 0.722).

**Table 2.** Performance comparison of adaptation methods across 10 held-out target tasks.

| Base model | Method | Params | NER | | RE | | QA | | NLI | | Summarization | | Avg. |
|---|---|---|---|---|---|---|---|---|---|---|---|---|---|
| | | | UW | Opioid | UW | Opioid | HEAD | MedB. | Sci | Rad | Rad | MeQ | |
| | | | F1 | F1 | F1 | F1 | Acc. | Acc. | F1 | F1 | R-L | R-L | |
| LLaMA 3.1 8B | Full FT | 100.0% | 0.832 | 0.877 | 0.796 | 0.858 | 0.584 | 0.627 | 0.836 | 0.784 | 0.364 | 0.435 | 0.699 |
| | LoRA | ~2.50% | 0.810 | 0.850 | 0.770 | 0.828 | 0.560 | 0.599 | 0.815 | 0.757 | 0.342 | 0.412 | 0.674 |
| | PT | <0.05% | 0.752 | 0.799 | 0.708 | 0.773 | 0.512 | 0.556 | 0.768 | 0.704 | 0.301 | 0.374 | 0.625 |
| | **MPT** | **<0.05%** | **0.824** | **0.868** | **0.789** | **0.850** | **0.573** | **0.616** | **0.829** | **0.773** | **0.357** | **0.432** | **0.691** |
| Meditron3 8B | Full FT | 100.0% | 0.861 | 0.904 | 0.823 | 0.882 | 0.603 | 0.651 | 0.849 | 0.811 | 0.379 | 0.458 | 0.722 |
| | LoRA | ~2.50% | 0.841 | 0.879 | 0.799 | 0.854 | 0.581 | 0.624 | 0.830 | 0.786 | 0.358 | 0.436 | 0.699 |
| | PT | <0.05% | 0.785 | 0.830 | 0.739 | 0.801 | 0.535 | 0.583 | 0.785 | 0.735 | 0.319 | 0.400 | 0.651 |
| | **MPT** | **<0.05%** | **0.857** | **0.895** | **0.817** | **0.875** | **0.595** | **0.643** | **0.843** | **0.803** | **0.372** | **0.454** | **0.715** |
| gpt-oss 20B | Full FT | 100.0% | 0.876 | 0.918 | 0.841 | 0.899 | 0.649 | 0.697 | 0.871 | 0.833 | 0.396 | 0.475 | 0.746 |
| | LoRA | ~2.50% | 0.858 | 0.895 | 0.819 | 0.873 | 0.629 | 0.673 | 0.853 | 0.810 | 0.377 | 0.454 | 0.724 |
| | PT | <0.05% | 0.804 | 0.848 | 0.761 | 0.822 | 0.585 | 0.634 | 0.810 | 0.761 | 0.340 | 0.420 | 0.679 |
| | **MPT** | **<0.05%** | **0.871** | **0.914** | **0.835** | **0.893** | **0.644** | **0.689** | **0.865** | **0.823** | **0.391** | **0.470** | **0.739** |

Full FT: full fine-tuning; PT: single-task soft prompt tuning from random initialization; MPT: multitask prompt distillation and decomposition (proposed). Params denotes the percentage of trainable parameters. UW: 2022 n2c2 University of Washington split; Opioid: UFHealth Opioid use dataset; HEAD: HEAD-QA dataset; MedB.: Medbullets dataset; Sci: SciNLI dataset; Rad: RadNLI dataset; MeQ: MeQSum dataset. R-L: ROUGE-L score; Avg.: mean score across all target tasks.

**Figure 2** shows the few-shot learning performance of MPT, LoRA, and PT across five clinical NLP task types at $k \in \{0, 1, 5, 10, 20\}$ labeled examples, averaged across three backbone models. The shaded bands denote standard deviation across 10 random draws of the k-shot training set. At $k = 0$, all three methods exhibit near-zero performance across all task types, confirming that zero-shot adaptation without any target supervision is insufficient for clinical NLP tasks. Across all task types and shot levels, MPT consistently outperforms both LoRA and PT, with the largest absolute margins observed at $k = 1$. At $k = 1$, MPT achieves 0.598, 0.431, 0.441, 0.541, and 0.158 for NER, RE, QA, NLI, and Summarization respectively, compared to 0.421, 0.274, 0.334, 0.388, and 0.094 for LoRA and 0.198, 0.109, 0.241, 0.224, and 0.041 for PT, with average improvements of 13.2% and 27.1% over LoRA and PT respectively. As $k$ increases from 1 to 20, all three methods improve synchronously across all task types. The gap between MPT and LoRA narrows most rapidly for NLI and NER as $k$ increases, while remaining largest for RE and Summarization at $k = 20$. At $k = 20$, MPT reaches 0.856, 0.758, 0.664, 0.831, and 0.394 for NER, RE, QA, NLI, and Summarization respectively. PT shows the most significant improvement curve from $k = 0$ to $k = 20$ across all task types but remains below both MPT and LoRA at every shot level. Shaded bands indicate that MPT also exhibits lower variance across random draws compared to LoRA and PT, most notably at $k = 1$ and $k = 5$.

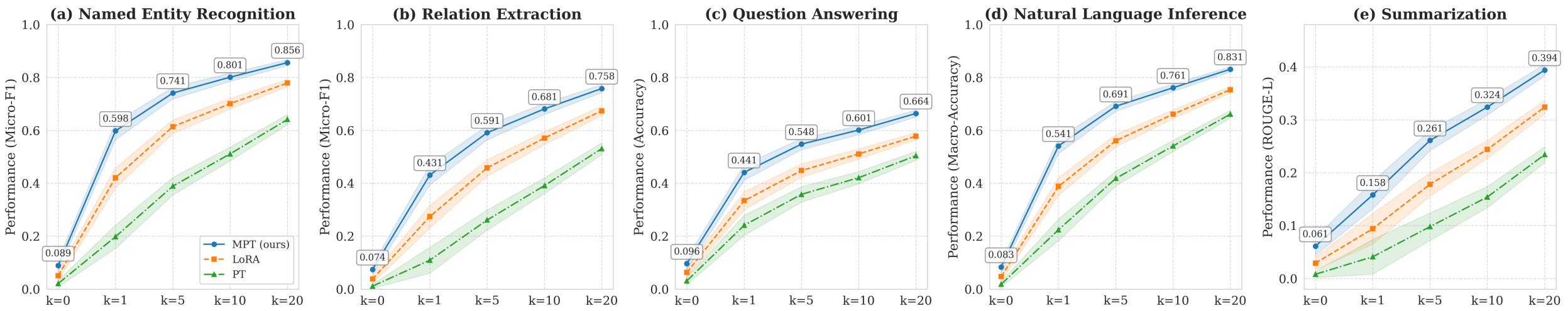

Figure 2. Few-shot learning performance of MPT, LoRA, and PT across five clinical NLP task types, averaged over three backbone models (LLaMA 3.1 8B, Meditron3 8B, and gpt-oss 20B).

## Discussion

Clinical NLP systems face persistent transfer learning bottlenecks across tasks, institutions, and disease domains, yet existing parameter-efficient adaptation methods either learn task-specific prompts independently without exploiting cross-task knowledge or rely on weight-space updates that impose significant storage and computing overhead at scale. This study addresses these challenges by adapting a multitask prompt distillation and decomposition framework to clinical NLP, demonstrating that a single shared meta-prompt learned from 21 diverse clinical source tasks can be efficiently adapted to unseen clinical target tasks across five task types, three backbone models, and both full-dataset and few-shot evaluation settings.

Our MPT framework achieves competitive performance with full fine-tuning and consistently outperforms the state-of-the-art PEFT method LoRA across all target tasks despite using three to four orders of magnitude fewer trainable parameters per target task. This result challenges the assumption that weight-space adaptation methods such as LoRA represent the optimal efficiency-performance tradeoff for clinical NLP. The practical implication is significant: a hospital system can maintain a single frozen backbone and a library of lightweight prompt vectors rather than storing and serving multiple LoRA-adapted model variants, substantially reducing deployment infrastructure requirements.

The substantial and consistent performance gap between MPT and single-task prompt tuning, averaging 6.1–6.6% across models under full dataset adaptation and widening further at low $k$, demonstrates that the shared meta-prompt encodes transferable clinical representations that single-task prompt tuning cannot acquire from target task data alone. This gap is largest on cross-institutional and cross-disease transfer tasks, where single-task PT shows the most significant performance drops. The few-shot results reveal that all three methods exhibit near-zero performance at $k = 0$, confirming that zero-shot clinical adaptation without any target supervision remains an open challenge regardless of the adaptation strategy. However, MPT recovers most rapidly with minimal supervision, achieving average improvements of 13.2% and 27.1% over LoRA and PT, respectively, at $k = 1$, demonstrating that the shared meta-prompt provides a strong initialization that requires very few labeled examples to activate task-specific adaptation. These properties make MPT particularly well-suited to low-resource clinical NLP deployment scenarios where obtaining annotated examples requires substantial clinical expert effort.

Meditron3 8B consistently outperforms LLaMA 3.1 8B across all methods and target tasks, with the largest gains on clinical structured prediction task (e.g., NER and RE). These differences are directly attributable to pretraining in the clinical domain. Notably, Meditron3 8B MPT outperforms LLaMA 3.1 8B full fine-tuning on average, demonstrating that clinical pretraining and multitask prompt transfer are complementary strategies whose combination can exceed a stronger general-domain model under full parameter adaptation. gpt-oss 20B achieves the highest overall performance across all tasks and methods, with the most significant advantages on QA tasks where model scale and MoE architecture contribute most to clinical reasoning. In comparison, its margin over Meditron3 8B is considerably smaller on structured prediction tasks where clinical pretraining partially compensates for the scale difference.

This study has limitations. While MPT is highly efficient, it is inherently bounded by the context window limits of soft continuous vectors. It relies heavily on the quality, diversity, and balanced sampling of the source distillation datasets. Future work should investigate dynamic routing of prompt segments based on the target task type, and evaluate how prompt distillation scales with the extended context windows of next-generation clinical foundation models. We focus on text-based clinical NLP tasks, and generalization to multimodal clinical data, including medical imaging, structured EHR records, and clinical time series, remains unexplored. We will explore multimodal prompt distillation frameworks that jointly encode textual and non-textual clinical data.

## Conclusion

This study presents a multitask prompt distillation and decomposition method for clinical NLP transfer learning. A single shared meta-prompt from diverse source tasks adapts to unseen clinical target tasks with fewer than 0.05% trainable parameters, consistently outperforming LoRA and single-task prompt tuning across five task types and three backbone models. MPT's strong few-shot performance makes it well-suited for low-resource clinical deployment. These results establish multitask prompt transfer as a practical foundation for scalable clinical NLP.

## Acknowledgments

We would like to thank the n2c2 challenge organizers for providing the annotated corpus. We also thank the developers and maintainers of the publicly available datasets used in this study, including BC5CDR, NCBI Disease, ChemProt, MedQA, MedMCQA, PubMedQA, emrQA, MedNLI, BioNLI, SDOH-NLI, NLI4CT, MIMIC-CXR, PubMed, and CHQ-Summ, whose open release made this benchmark possible. We acknowledge the support from the Cancer Informatics Shared Resource in the UF Health Cancer Center. We would like to thank the UF Research Computing team for providing computing power through the UF HiPerGator-AI cluster.

## Funding statement

This study was partially supported by grants from the Patient-Centered Outcomes Research Institute® (PCORI®) Award ME-2023C3-35934, the PARADIGM program awarded by the Advanced Research Projects Agency for Health (ARPA-H), National Institute on Aging U24AG098157, National Institute of Allergy and Infectious Diseases, NIAID R01AI172875, National Heart, Lung, and Blood Institute, R01HL169277, R01HL176844, National Institute on Drug Abuse, NIDA R01DA057886, R01DA063631, and the UF Clinical and Translational Science Institute. The content is solely the responsibility of the authors and does not necessarily represent the official views of the funding institutions.